%% file: main.tex
\documentclass{article}

    \PassOptionsToPackage{numbers, compress,sort}{natbib}

\usepackage[preprint]{neurips_2025}

\usepackage{amsmath} 
\usepackage[utf8]{inputenc} 
\usepackage[T1]{fontenc}    
\usepackage{url}            
\usepackage{booktabs}       
\usepackage{amsfonts}       
\usepackage{nicefrac}       
\usepackage{microtype}      
\usepackage[table,dvipsnames,x11names]{xcolor}         
\input{preamble}
\usepackage{multirow}
\usepackage{graphics}
\usepackage{graphicx}
\usepackage{soul}
\usepackage[export]{adjustbox}
\usepackage[font=small,tableposition=top]{caption}
\usepackage{pifont}
\usepackage[most]{tcolorbox}
\usepackage{listings}
\definecolor{mydarkgreen}{HTML}{009901}
\usepackage{makecell}
\usepackage{wrapfig}
\usepackage{booktabs}
\usepackage{amssymb}
\usepackage{hyperref}       

\colorlet{punct}{red!60!black}
\definecolor{background}{HTML}{EEEEEE}
\definecolor{delim}{RGB}{20,105,176}
\colorlet{numb}{magenta!60!black}
\lstdefinelanguage{json}{
    basicstyle=\tiny\ttfamily,
    numbers=left,
    numberstyle=\scriptsize,
    stepnumber=1,
    numbersep=8pt,
    showstringspaces=false,
    breaklines=true,
    backgroundcolor=\color{background},
    literate=
     *{0}{{{\color{numb}0}}}{1}
      {1}{{{\color{numb}1}}}{1}
      {2}{{{\color{numb}2}}}{1}
      {3}{{{\color{numb}3}}}{1}
      {4}{{{\color{numb}4}}}{1}
      {5}{{{\color{numb}5}}}{1}
      {6}{{{\color{numb}6}}}{1}
      {7}{{{\color{numb}7}}}{1}
      {8}{{{\color{numb}8}}}{1}
      {9}{{{\color{numb}9}}}{1}
      {:}{{{\color{punct}{:}}}}{1}
      {,}{{{\color{punct}{,}}}}{1}
      {\{}{{{\color{delim}{\{}}}}{1}
      {\}}{{{\color{delim}{\}}}}}{1}
      {[}{{{\color{delim}{[}}}}{1}
      {]}{{{\color{delim}{]}}}}{1},
}

\title{Ego-R1: Chain-of-Tool-Thought for \\ Ultra-Long Egocentric Video Reasoning}

\author{
Shulin Tian$^{1,2*}$ \quad Ruiqi Wang$^{1,3*}$ \\
\textbf{Hongming Guo}$^4$ \quad \textbf{Penghao Wu}$^{1}$  \quad \textbf{Yuhao Dong}$^{1}$  \quad \textbf{Xiuying Wang}$^{1}$\\
\textbf{Jingkang Yang}$^{1}$  \quad \textbf{Hao Zhang}$^3$ \quad \textbf{Hongyuan Zhu}$^2$ \quad \textbf{Ziwei Liu}$^{1}$ \\
$^1$S-Lab, Nanyang Technological University \quad $^2$A*STAR, Singapore \\
$^3$Simon Fraser University \quad $^4$Shanghai AI Lab
}
\begin{document}
\maketitle
\vspace{-3em}
\begin{center}
    \includegraphics[width=\linewidth]{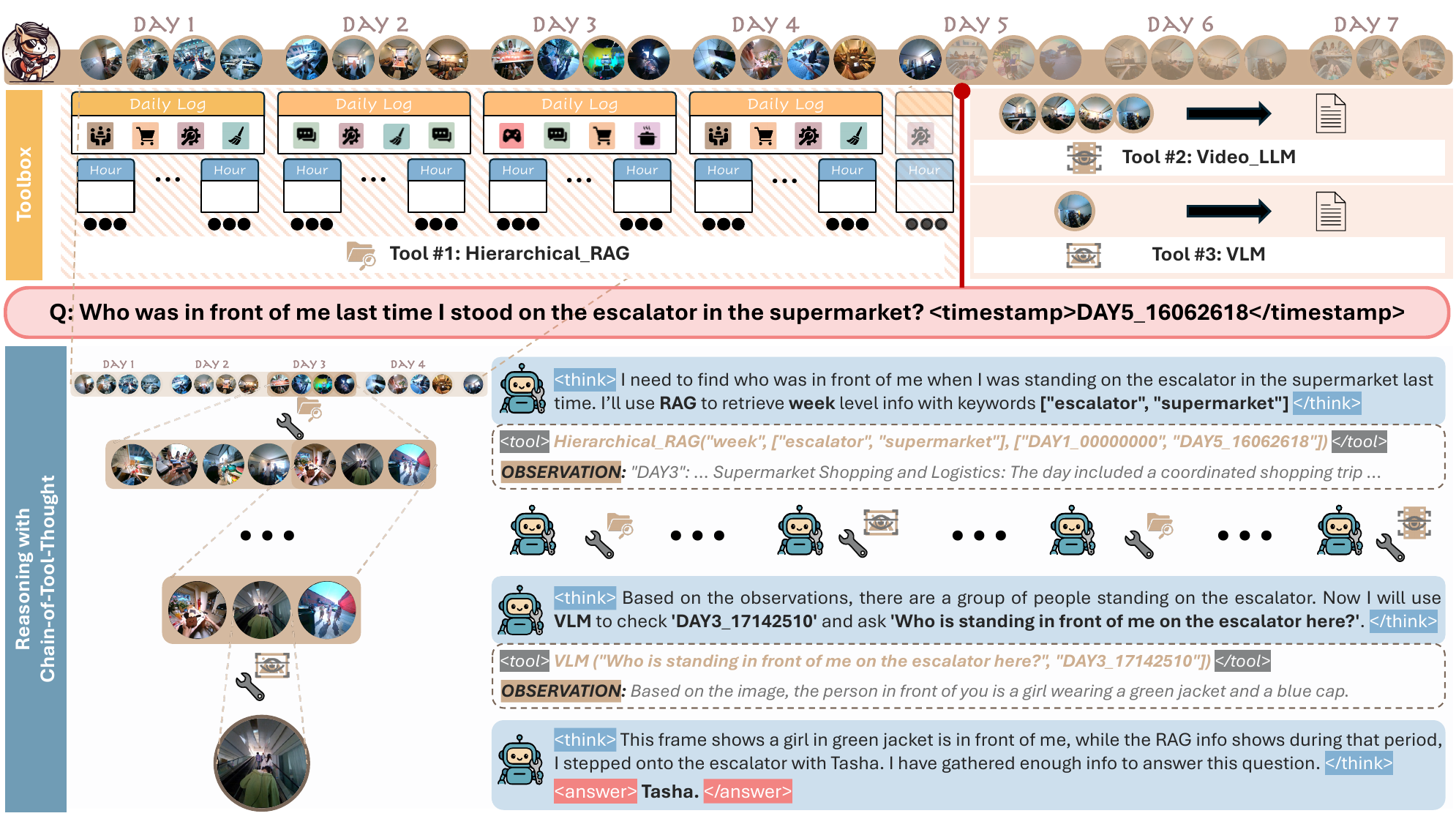}
    \vspace{-1.5em}
    \captionof{figure}{\textbf{Overview of Ego-R1.} 
    In this figure, we demonstrate how the Ego-R1 Agent orchestrates specialized tools (e.g., Hierarchical\_RAG, Video LLM, and VLM) to answer the question step-by-step, based on the observations and previous actions.
    The system effectively answers questions that require careful searching within ultra-long videos and precise analysis of frame details.
    }
    \label{fig:teaser}
  \end{center}

\vspace{-2pt}
\begin{abstract}
\vspace{-6pt}
  We introduce \textbf{Ego-R1}, a novel framework for reasoning over \rz{\emph{ultra-long} (i.e., in days and weeks)} egocentric videos, which leverages a structured \textbf{Chain-of-Tool-Thought (CoTT)} process, orchestrated by an \textbf{\ourmodel{}} trained via reinforcement learning \rz{(RL)}.
  Inspired by human problem-solving strategies, CoTT decomposes complex reasoning into modular steps, with \rz{the RL agent invoking specific tools, one per step, to iteratively and collaboratively answer sub-questions tackling such tasks as temporal retrieval and multi-modal understanding.}
  \rz{We design a two-stage training paradigm involving supervised finetuning (SFT) of a pretrained language model using CoTT data and RL to enable our agent to dynamically propose step-by-step tools for long-range reasoning.
  To facilitate training, we construct a dataset called \textbf{Ego-R1 Data}, which consists of \textbf{Ego-CoTT-25K} for SFT and \textbf{Ego-QA-4.4K} for RL.} Furthermore, our Ego-R1 agent is evaluated on a newly curated week-long video QA benchmark,\textbf{~\ourbenchmark}, which contains human-verified QA pairs from hybrid sources. \rz{Extensive results demonstrate that the dynamic, tool-augmented chain-of-thought reasoning by our \ourmodel{} can effectively tackle the unique challenges of understanding ultra-long egocentric videos, significantly extending the time coverage from few hours to a week.}
\end{abstract}

\input{secs/1_introduction}
\input{secs/2_related}

\input{secs/3_task}
\input{secs/4_method}

\input{secs/5_experiments}

\input{secs/6_conclusion}
\newpage

\bibliographystyle{plain}
\bibliography{ref}
\newpage

\input{secs/X_appendix}

\end{document}

%% file: preamble.tex
\newcommand{\ourmethod}{Ego-R1}
\newcommand{\ourmodel}{Ego-R1 Agent}
\newcommand{\ourdata}{Ego-R1 Data}
\newcommand{\ourbenchmark}{Ego-R1 Bench}
\newcommand{\toolrag}[0]{\includegraphics[scale=0.5,valign=c]{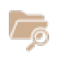}}
\newcommand{\toolvideollm}[0]{\includegraphics[scale=0.5,valign=c]{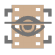}}
\newcommand{\toolvlm}[0]{\includegraphics[scale=0.5,valign=c]{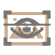}}
\newcommand{\cmark}{\textcolor[rgb]{0.13, 0.55, 0.13}{\ding{51}}}%
\newcommand{\xmark}{\textcolor[rgb]{0.5, 0, 0}{\ding{55}}}%

\newcommand{\rqw}[1]{\textcolor{blue}{#1}}
\newcommand{\rz}[1]{\textcolor{black}{#1}}

\definecolor{darkblue}{RGB}{0, 0, 139} 
\newcommand{\hy}[1]{\textcolor{darkblue}{#1}}

%% file: secs/1_introduction.tex
\section{Introduction}
\label{sec:intro}

Egocentric videos, which capture human daily lives from a first-person perspective, are inherently long - often spanning hours to days or even weeks~\cite{yang2025egolifeegocentriclifeassistant}. Understanding these videos is crucial for supporting practical tasks \rz{such as} memory recall, multi-step activity tracking, and goal monitoring~\cite{grauman2024ego,mangalam2023egoschema,chen2024groundedmultihopvideoqalongform}. But the ensuing problem poses significant challenges due to the video length, multi-modality, and the need for long-horizon reasoning across diverse temporal contexts \rz{and dependencies}.

Recent advances in multimodal long-context modeling have led to promising progress, extending video understanding capabilities from minutes to hours~\cite{zhang2024longva,chen2024longvila,zhang2024llavanext-video,li2024videochatchatcentricvideounderstanding}. However, these models still face significant computational challenges and scale poorly when applied to videos of extended durations, such as those spanning a day \rz{or longer. To this end,} prior works have proposed token compression~\cite{Song_2024_CVPR,shen2024longvuspatiotemporaladaptivecompression,weng2024longvlmefficientlongvideo,shu2024videoxlextralongvisionlanguage,jiang2025tokenefficientlongvideounderstanding} or sampling-based strategies that reframe video understanding as a temporal retrieval task~\cite{qu2025doesvisionlanguagemodellost,ye2025re}. Nevertheless, these approaches risk missing key events due to the lossy representations or incomplete temporal localization. Another line of works, commonly referred to as video agents, leverages external language models as high-level \rz{control and reasoning entities} to call specialized vision modules\rz{/tools} for video reasoning~\cite{wang2024videoagent,ye2025re,zhi2025videoagent2enhancingllmbasedagent}. While \rz{allowing more flexible and more granular perception, these approaches still rely on predefined reasoning pipelines or fixed-order tool invocations, limiting the video lengths they can handle, i.e., up to hour-long.}

To address these limitations, we propose \textbf{\ourmethod{}}, a novel framework that leverages \rz{fine-tuned large language models (LLMs) and reinforcement learning (RL)} for {\em dynamic\/} tool-driven reasoning of \rz{{\em ultra-long\/} (i.e., in days and weeks) egocentric videos}. \rz{The key distinction from prior video agents~\cite{wang2024videoagent,ye2025re,zhi2025videoagent2enhancingllmbasedagent} designed for long-form video understanding is the dynamic tool calling of our \textbf{\ourmodel{}}, which iteratively processes both visual information and contexts to select and execute specialized perception tools {\em on demand\/}, based solely on previously observed content and thought to preceding sub-questions. We call such a video understanding paradigm \textbf{Chain-of-Tool-Thought (CoTT)} reasoning.}
Furthermore, unlike traditional methods that either feed the entire video to the model or select a subset of the frames, Ego-R1 utilizes a structured toolkit for perception which consist of three core modules designed specifically to facilitate efficient temporal retrieval and detailed visual comprehension. 
For retrieval, \textit{Hierarchical Retrieval-Augmented Generation (H-RAG)} extracts timestamped, question-relevant information in the language space. For visual analysis, a specialized \textit{Video-LLM} interprets localized visual contexts, while a general-purpose \textit{Vision-Language Model (VLM)} extracts fine-grained visual details. Coordinated by an orchestrating LLM \rz{trained through RL}, Ego-R1 enables scalable, step-by-step compositional reasoning \rz{over ultra-long} videos. The modular design of our framework enables easy integration with a wide range of state-of-the-art visual understanding models, allowing the visual perception components, i.e., the Video-LLM and VLM, to seamlessly integrate into our framework.

To facilitate the training of \ourmethod{}, \rz{which consists of a supervised fine-tuning (SFT) stage and an RL stage,} we construct \textbf{\ourdata}, a comprehensive hybrid-source dataset consists of 25K CoTT reasoning traces and 4.4K annotated question-answer (QA) instances \rz{to support SFT of a pretrained LLM and RL training of our Ego-R1 agent, respectively}. Each task within the dataset requires reasoning over substantial temporal spans, with an average of 7.42 tool-calling steps per task. Additionally, we introduce~\textbf{\ourbenchmark}, a carefully curated evaluation framework \rz{consisting of week-long egocentric videos} that combine human-annotated and post-verified synthetic data, designed specifically to assess long-horizon reasoning capabilities in \rz{the egocentric setting}.


\rz{
Extensive experiments across diverse long-video benchmarks demonstrate that the dynamic, tool-augmented chain-of-thought reasoning by our Ego-R1 Agent can effectively tackle the unique challenges of understanding ultra-long egocentric videos, significantly extending the time coverage from few hours to a week.
We also perform ablation studies to replace the visual modules in~\ourmethod{} to showcase that our framework is customized to integrate current MLLMs scope, validating our method's robustness and generalization.  
At last, while we focus on egocentric long videos in this work, we show that our framework generalizes well in the exocentric setting as well.
}

\input{tabs/method-comparison}

%% file: tabs/method-comparison.tex
\begin{table*}[t]
\centering
\caption{\textbf{Comparison between \ourmethod{} and other frameworks.} \ourmethod{} develops an agentic tool-calling schema that enables interpretable reasoning over ultra-long videos while preserving critical temporal information.}
\vspace{-0.5em}
\label{tab:comparison}
\footnotesize
\resizebox{1.0\textwidth}{!}{%
\begin{tabular}{lccccc}
\toprule
\textbf{Method} & \textbf{Ultra Long}  & \textbf{Information Retention} & \textbf{Interpretability} & \textbf{Temporal Awareness} & \textbf{Adaptability}\\
\midrule
LongVA~\cite{zhang2024longva} & \xmark & \xmark & \xmark & \xmark & \xmark \\
LLaVA-Video~\cite{zhang2024llavanext-video} & \xmark & \xmark & \xmark & \xmark & \xmark\\
$T^*$~\cite{ye2025re} & \xmark & \cmark & \xmark & \cmark & \cmark \\
Video-RAG\cite{luo2024video} & \xmark & \cmark & \xmark & \cmark & \cmark \\
VideoAgent~\cite{wang2024videoagent} & \xmark & \cmark & \xmark & \cmark & \cmark \\
Video-R1~\cite{feng2025video} & \xmark & \xmark  & \cmark & \xmark & \xmark \\
\midrule
\ourmethod & \cmark & \cmark & \cmark & \cmark & \cmark \\
\bottomrule
\end{tabular}%
}
\vspace{-5pt}
\end{table*}

%% file: secs/2_related.tex
\section{Related Work}
\noindent\textbf{Egocentric long video understanding.}
Existing large-scale egocentric datasets such as Ego4D~\cite{grauman2022ego4d}, EgoExo4D~\cite{grauman2024ego}, Epic-Kitchens~\cite{damen2018scaling}, and HD-Epic~\cite{perrett2025hd} have established comprehensive benchmarks~\cite{mangalam2023egoschema,cheng2024egothink,chen2023egoplan} focused on temporal understanding of daily activities, object interactions, and episodic memory tasks~\cite{song2023ego4d,li2025egotom,di2024grounded,tang2023egotracks,rodin2024action,goletto2024amego}. While these benchmarks typically span only minutes, recent extensions have reached hours~\cite{chandrasegaran2024hourvideo,ye2025re} but multi-personal interactions and cross-day behavioral patterns remain unexplored. Recently, EgoLife~\cite{yang2025egolifeegocentriclifeassistant} provides a week-long egocentric dataset; however, its question-answering tasks remain vanilla, lacking requirements for deep visual reasoning. Our benchmark addresses these limitations with more challenging tasks requiring sophisticated reasoning about visual details across diverse scenarios.


While egocentric datasets and benchmarks continue to expand in temporal scope, methods specifically designed for egocentric long video understanding remain absent. 
As shown in Table~\ref{tab:comparison}, existing approaches face critical limitations: 
proprietary models~\cite{achiam2023gpt, team2024gemini} and some MLLMs~\cite{li2024llava, bai2025qwen2} usually process videos as unified inputs, which becomes prohibitively token-intensive for hour-long videos; general frame sampling approaches~\cite{zhang2024longva, zhang2024llavanext-video, wang2025internvideo2,liu2024oryx,liu2025ola} cannot guarantee question-relevant frames selection; and sophisticated video agents~\cite{wang2024videoagent,ye2025re,Song_2024_CVPR, wang2023lifelongmemory, wang2024videotree} analyze frames in isolation, missing narrative structure and temporal dynamics. 
Though RAG shows promising direction for long video understanding~\cite{luo2024video, xu2024retrieval}, existing approaches often lack contextual specificity for multi-day egocentric videos, where personal routines and social dynamics evolve over time. To address this challenge, our \ourmethod{} implements multi-step reasoning upon a hierarchical RAG paradigm, enabling comprehensive understanding of evolving contexts beyond previous single thinking step approach Video-R1~\cite{feng2025video}. A detailed qualitative results comparison is shown in Fig.~\ref{fig:supp-qual}.

\noindent\textbf{Multimodal agentic tool-use.}
Agentic systems with Tool-Integrated Reasoning (TIR) effectively enhance LLMs' complex problem-solving and reasoning capabilities~\cite{parisi2022talm, yao2023react}, particularly in mathematical domains~\cite{gou2023tora, wang2023mathcoder,zhou2023solving, yang2024qwen2} through search engines~\cite{jin2025search,zheng2025deepresearcher} and code interpreters~\cite{yao2025fans,liao2024mario,yang2024octopus}. For training paradigms in tool-integrated learning, RL has emerged as a promising approach offering more scalable and generalizable tool utilization strategies~\cite{qian2025toolrl,li2025torl, feng2025retool,wang2025otc}, compared to traditional SFT~\cite{schick2023toolformer,qin2023toolllm}. 
Recent research has extended tool-augmented foundation models to multimodal domains, exploring the integration of diverse tool-use for visual reasoning tasks~\cite{ke2025dwim, deng2025openvlthinker, su2025openthinkimglearningthinkimages, maaz2024videochatgptdetailedvideounderstanding,zhi2025videoagent2enhancingllmbasedagent, ma2024tacolearningmultimodalaction}. These initial efforts leverage specialized visual perception modules~\cite{wang2024videoagent, fan2025videoagent}, to enhance grounded and context-aware reasoning in complex visual environments~\cite{chen2024longvila,liu2024coarse}. Coinciding with OpenAI's o3~\cite{openai2025o3}, \ourmodel{} employs dynamic tool-calling mechanisms, enabling multi-step reasoning and contextual tool selection, determining the appropriate tool for optimal problem-solving. 

\noindent\textbf{CoT reasoning.} 
Chain-of-Thought (CoT) reasoning~\cite{wei2022chain} has emerged as a fundamental mechanism to enhance the reasoning capabilities of both LLM and VLM~\cite{xu2024llava, thawakar2025llamav, wang2025resanything, wu2025grounded,liu2024chain}.
RL-based reasoning approaches further require high-quality CoT samples to advance multimodal reasoning capabilities~\cite{huang2025vision,yang2025r1,dong2024insight,zhang2025r1}. 
However, existing datasets lack adequate, high-quality CoT annotations for long video understanding tasks. 
To fill this gap, we introduce Ego-CoTT-25K, featuring CoT reasoning with dynamic tool-calling capabilities.

%% file: secs/3_task.tex
\section{Egocentric Long Video Reasoning via Dynamic Tool-Calling}
\label{sec:task}

The egocentric long-video reasoning task represents a crucial frontier beyond understanding, as first-person perspectives capture complex, temporally interdependent human behaviors over ultra-long durations. Actions that occur many hours or even days apart may be guided by consistent personal strategies and habits; thus, correctly answering a query often relies on recognizing enduring human traits and linking them to cues dispersed across the entire timeline. 
This requires the models to therefore maintain long-range temporal dependencies, identify subtle evidence in earlier segments, and reason about the actor's underlying preferences to generate dynamic context-aware solutions. 
Although recent MLLMs demonstrate promising performance in general video understanding, they still struggle in answering questions in truly long-context videos with extended temporal relationships. This underscores the importance of egocentric long-video reasoning as a fundamental challenge for multimodal systems. 
In this section, we introduce Ego-R1, a novel framework that unifies visual content comprehension and contextual reasoning by combining chain-of-thought prompting with dynamic tool calling. We provide a formal task definition in Section~\ref{subsec: 3-1}, followed by a comprehensive presentation of our specialized toolkit architecture designed for dynamic tool call in Section~\ref{subsec: 3-2}.

\subsection{Egocentric Long Video Reasoning Tasks}
\label{subsec: 3-1}
Compared to general exocentric videos, egocentric videos offer continuous, context-rich recordings from a first-person perspective, naturally documenting extensive temporal experiences including daily routines, social interactions, and object manipulations. This unique viewpoint requires sophisticated high-order inference to interpret actions, intentions, and contexts across substantial temporal spans, making it require reasoning models with strong temporal understanding and contextual integration capabilities. This necessitates a flexible reasoning framework that dynamically processes both visual information and contextual details through an intelligent tool-calling mechanism, determining which analytical approaches are most relevant for comprehending complex temporal narratives spanning multiple days of recorded experience.

In our task, we provide egocentric video spanning several days alongside questions posed at a specific query time. The system analyzes all preceding video content to generate accurate responses, simulating human temporal reasoning in real-life scenarios. This tool-based approach enables multimodal reasoning by leveraging contextual information across extended periods, requiring the system to choose optimal tools during the thinking process to effectively integrate perception, memory, and action when generating responses based solely on previously observed content.

\subsection{Dynamic Tool-Calling}
\label{subsec: 3-2}



Current MLLMs struggle with extended egocentric content due to limited context windows, inadequate temporal understanding, and insufficient structured reasoning capabilities, preventing effective analysis of long-duration egocentric videos containing sparse events that require multi-step, context-aware interpretation. To address the inherent difficulty posed by the overly long context of long-form egocentric video reasoning, we adopt a \textit{dynamic tool-calling} framework that empowers the LLM rather than an MLLM to invoke specialized perception tools on demand. Our approach enables the LLM to actively decompose complex queries, selectively retrieve relevant segments, and iteratively perform stepwise reasoning grounded in video observations. This modular design overcomes the context-length bottleneck of MLLMs while enabling the fine-grained, multi-turn reasoning essential for practical egocentric video understanding. Our framework leverages three complementary tools - one text-based and two visual-based - each addressing distinct temporal and perceptual dimensions of egocentric understanding. The text-based hierarchical RAG system handles longer temporal information retrieval, while the visual-based tools (Video-LLM and VLM) perform detailed visual analysis at different visual granularities.

\texttt{h-rag}\toolrag:
Our hierarchical system efficiently localizes relevant temporal information from the memory bank. Videos are first segmented into 30-second clips, each summarized via a video captioning model and temporally aligned with the ASR results as clip logs. These clip logs are hierarchically aggregated through a \textit{bottom-up} generation process into multi-level granularity, creating comprehensive temporal summaries. The hierarchical structure facilitates effective \textit{top-down} inference to locate and retrieve logs of relevant video segments, thus reducing computational load while preserving accuracy and temporal coherence across long egocentric videos spanning days. The system accepts specific search parameters, including temporal granularity, keywords, and time ranges for retrieval, returning the most relevant observations that match the query constraints. 


\texttt{video-llm}~\toolvideollm:
Our \texttt{video-llm} is a short-horizon visual-perception module that operates on local temporal windows ranging from a few seconds up to ten minutes. We sample each clip within the proposed time range at 1 FPS, keeping the input size compatible with modern multimodal language models and thus maintaining broad architectural flexibility. Given a question and its corresponding video segment, the tool correlates visual content with temporal context to produce detailed observations that capture dynamic interactions and sequential events, and, when possible, directly answers the query for the specified time range.

{\texttt{vlm}}~\toolvlm: 
This general-purpose \texttt{vlm} operates at the finest temporal granularity, analyzing individual frames to extract high-resolution details like text on packaging, object attributes or specific visual elements missed in broader video analysis. It augments the temporal reasoning of \texttt{video-llm} with precise visual evidence for comprehensive egocentric understanding. 

%% file: secs/4_method.tex
\section{Ego-R1 Data: Chain-of-Tool-Thought (CoTT) for Video Reasoning}
\label{sec:method}
\begin{figure}
    \centering
    \includegraphics[width=\linewidth]{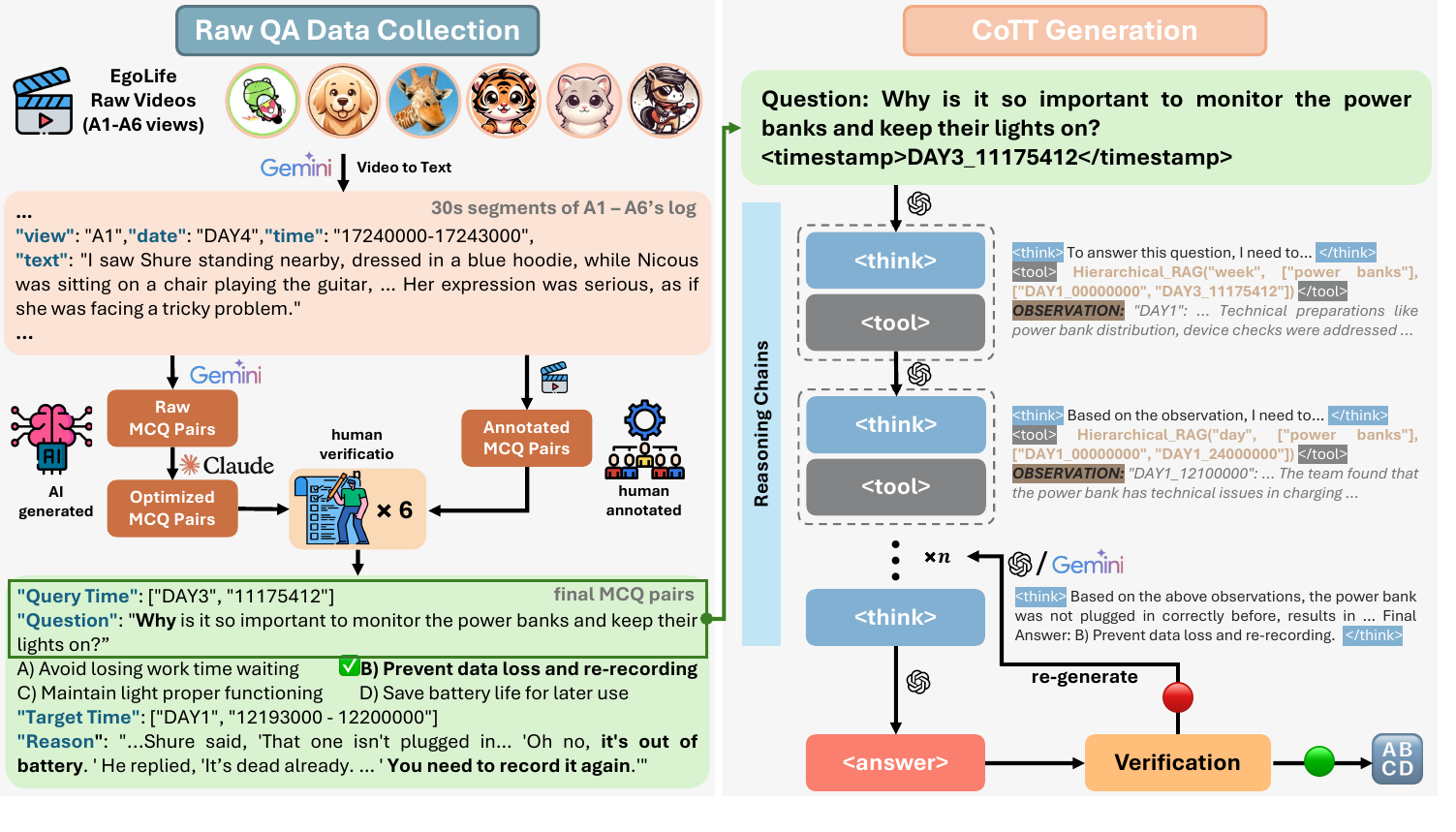}
    \vspace{-2em}
    \caption{\textbf{Data generation pipeline of the \ourdata.} We first obtained raw QA pairs from both AI-generated and human-annotated sources based on 6 raw videos collected from 6 participants and the corresponding log. The verified and processed Multiple Choice Questions (MCQs) serve as the foundation of the \ourdata{} (left). We take questions without answers for Chain-of-Tool-Thought (CoTT) generation, which involves creating reasoning chains that include explicit thinking steps and dynamic tool-calling sequences (right).}
    \label{fig:data-gen}
    \vspace{-1.5em}
\end{figure}

To unleash the reasoning capabilities of LLM under the CoT prompting paradigm and to enable dynamic tool selection conditioned on current observations and past actions, we introduce \ourdata{}, a dataset designed to enable agentic tool-use with Chain-of-Tool-Thought (CoTT) reasoning chains. Figure ~\ref{fig:data-gen} illustrates the data generation pipeline of the \ourdata, including raw QA data collection and CoTT generation. 
In this section, we define the structure of CoTT in Section~\ref{ssec:cott-def}, and provide details of \ourdata{} generation in Section~\ref{ssec:cott-data-gen}.



%
\subsection{Chain-of-Tool-Thought (CoTT)}
\label{ssec:cott-def}
Our goal is to generate synthetic CoTT data and use it to train multi-turn tool-use language models.
We define a CoTT data $C$ as a sequence of steps $S_i$, where each step consists of a thought $T^{\mathrm{th}}_i$, a tool $T^{\mathrm{to}}_i$, and an observation $o_i$. A CoTT trajectory is defined as follows:
\begin{equation}
    C = (S^0, S^1, \dots, S^n),  
    \quad S^i = \bigl(\,T^{\mathrm{th}}_i,\,T^{\mathrm{to}}_i,\,o_i\bigr)
\end{equation}
where $C$ is a sequence of $n$ reasoning steps. At each step $i$, the agent will generate a thought $T^{\mathrm{th}}_i$ and a tool call $T^{\mathrm{to}}_i$ based on all the previous steps' observations $\{o_0, o_1, \dots, o_{i-1}\}$ and the query $q$.

To formalize this reasoning process, we define two essential components that characterize how the agent operates: the action space, which specifies the available tools the agent can utilize, and the observation space, which captures the structured outputs returned from tool executions.

\noindent\textbf{Action space.} We define the action space $\mathcal{A} = {F_j}$ as a union of available tools to be used during reasoning. We use the three fundamental tools defined in Section~\ref{subsec: 3-2}: 1)~\texttt{h-rag} for text-based long-range temporal retrieval, 2)~\texttt{video-llm} for short-range video understanding, and 3)~\texttt{vlm} for framewise image understanding, plus an auxiliary \texttt{terminate} tool for data generation only.
The \texttt{h-rag} tool retrieves relevant information from the current-view knowledge base by querying specified keywords within a target time window. By projecting long videos into a semantically and temporally structured language space, it rapidly pinpoints the approximate temporal interval of an event while summarizing sparse visual cues into a concise textual summary.
The \texttt{video-llm} tool analyses short video segments specified by a query and an associated time window, providing detailed interpretations of local visual–temporal content.
The \texttt{vlm} tool performs image-level analysis on a single frame selected by timestamp and query, providing precise, frame-specific visual details.

\noindent\textbf{Observation space.} At each reasoning step $i$, the agent receives an observation $o_i = \bigl(o_i^{\text{rag}},\,o_i^{\text{vid}},\,o_i^{\text{vlm}}\bigr)\;\in\;\mathcal{O}$, where each component $o_i^{\text{rag}}$, $o_i^{\text{vid}}$, $o_i^{\text{vlm}}$ represents the output of corresponding tool \texttt{rag}, \texttt{video-llm}, and \texttt{vlm}. The observation space $O = \{O^0, O^1, ..., O^n\}$ encompasses the collection of all tool outputs. Each tool call executes via the parsed arguments, producing observations that guide subsequent reasoning steps.

\vspace{-1em}
\subsection{Data Generation}
\label{ssec:cott-data-gen}
\vspace{-0.5em}
We carefully curate \ourdata, comprising 4.4K annotated question-answer pairs sourced from over 500 hours of egocentric videos recorded across six distinct first-person perspectives. We select 2.9K high-quality questions for CoTT generation. For each selected QA pair, we construct a CoTT trace that decomposes the reasoning process into interpretable steps, yielding an average of 7.42 tool calls per task. In total, 25K CoTT traces are generated, and subsequently used during the SFT stage to train our multi-turn tool-use language model.

\noindent\textbf{Ego-QA-4.4K.} Long-form egocentric videos are hard to collect in nature. Following the dataset construction pipeline of EgoLifeQA~\cite{yang2025egolifeegocentriclifeassistant}, we collected 2.9K high-quality human-annotated data from 6 videos with distinct viewpoints.  To expand the dataset scale, we employ proprietary models to analyze Automatic Speech Recognition (ASR) transcripts with video captioning outputs from the 30-second segments. These textual logs were combined and examined across various temporal granularities, spanning single or multiple days, to generate candidate questions with answers. Human annotators subsequently selected and cross-validated those QA pairs using Fleiss' kappa~\cite{fleiss1973equivalence}, refining each query and its ground-truth answer according to a unified criteria of rationale coherence, importance, relevance, and difficulty level.
In total, \ourdata{} comprises $4.4$K question-answer pairs from both human-labeled and synthetic data sources.

\noindent\textbf{Ego-CoTT-25K.} We develop a systematic CoTT generation system to automatically generate CoTT data based on the selected question-answer pairs. By leveraging proprietary LLMs with longer context windows and stronger instruction-following capabilities, we enable the automatic generation of comprehensive reasoning chains that would otherwise be challenging to produce manually.
In the CoTT generation system, each tool is exposed to the model as an executable function whose signature and semantics are implicitly embedded in the system. This design, paired with a textual system prompt (Table \ref{tab:supp-gen-prompt}), prevents parsing errors during execution.
The prompt also encodes the current viewpoint identity and enumerates the available tools. Given an input question $q$, the model iteratively generates reasoning steps $S_i = (T^{th}_i, T^{to}_i)$, where $T^{th}_i$ denotes the thought and $T^{to}_i$ denotes the corresponding tool call with fully specified arguments (e.g., time ranges, keywords, sub-questions). All the proposed arguments are validated by a pre-verification module to ensure syntactic correctness. Once a call is emitted, its name and arguments are extracted via special tokens and dispatched to an external server for execution. The returned observation is then fed back to the model, guiding the next step and enabling dynamic, multi-turn tool use for egocentric long-video reasoning.

\section{Ego-R1 Agent: Towards Tools Integrated Video Understanding Agent}
\begin{figure}
    \centering
    \includegraphics[width=1.\linewidth]{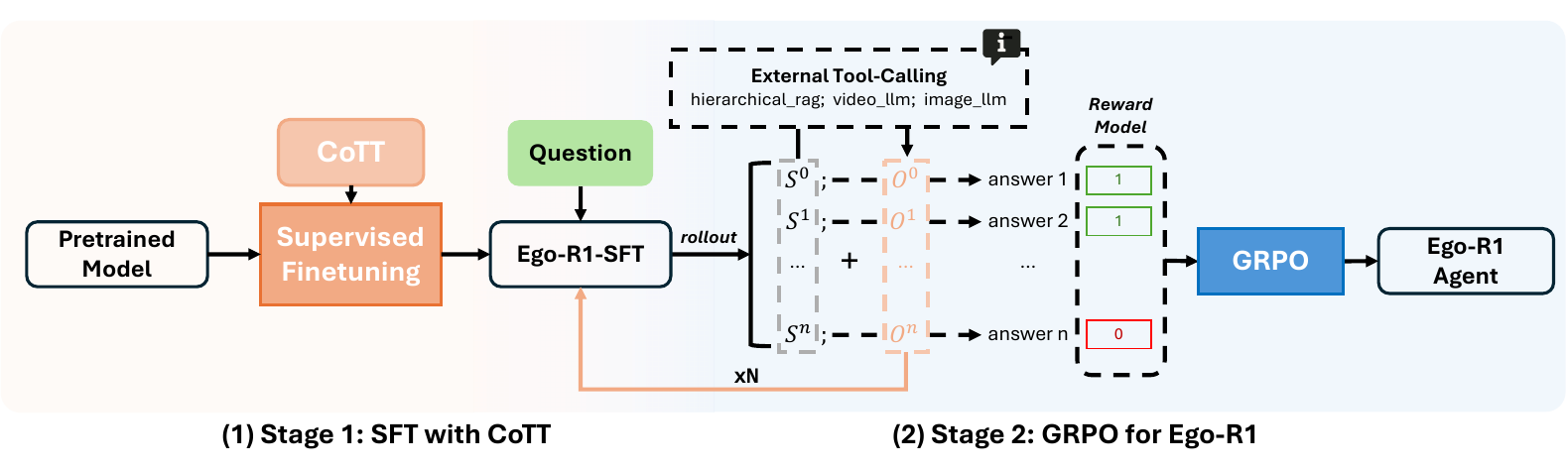}
    \vspace{-10pt}
    \caption{\textbf{Overview of the two-stage training strategies in Ego-R1.} Ego-R1 employs a two-stage training approach: Stage 1 utilizes supervised fine-tuning with CoTT data to establish structured tool-calling capabilities, while Stage 2 applies multi-turn reinforcement learning with rule-based rewards to optimize iterative reasoning and tool execution across diverse question types.}
    \label{fig:training}
\end{figure}
Our goal is to train a language model capable of performing long-form video reasoning via a structured long-chain reasoning schema that automatically invokes multi-turn tool calls to collaboratively solve the problem. Inspired by the recent post-training techniques~\cite{chu2025sftmemorizesrlgeneralizes}, we design our training framework with a two-stage strategy, with an illustration in Fig.~\ref{fig:training}. 

\subsection{Stage 1: Supervised fine-tuning (SFT)}

In the first stage, we perform SFT on a pretrained language model using the synthetic CoTT dataset. This "cold-start" initialization equips the model with the foundational ability to produce correctly formatted tool calls as prescribed by the CoTT reasoning schema. The CoTT data, presented in a structured, multi-turn conversational format, simulates realistic stepwise tool interactions, explicitly combining natural language reasoning with structured tool invocation. Each step in the reasoning trajectory consists of a thought enclosed within the special token \texttt{<think>...</think>}, followed by either a proposed tool call, enclosed within \texttt{<tool>...</tool>}, or an answer, enclosed with in \texttt{<answer>...</answer>}. The tool call is automatically parsed and executed by an external environment, which then returns an observation. This observation is formatted and fed back into the model as part of the input for the next reasoning step.
After fine-tuning, the resulting \emph{Ego-R1-SFT} model reliably produces well-formed tool calls and coherent step-by-step reasoning, laying the groundwork for subsequent reinforcement learning stage.

\subsection{Stage 2: Reinforcement learning (RL)}
To further improve the multi-turn tool-calling capabilities of our fine-tuned \emph{Ego-R1-SFT} model, we adopt \textit{Gradient-Regularized Policy Optimization} (GRPO)~\cite{shao2024deepseekmath} to train the model. GRPO optimizes the model to maximize the expected final task reward while regularizing the variance of policy gradients across reasoning steps to encourage stable and coherent decision-making. Specifically, we define the GRPO objective as follows:
\begin{align*}
\mathcal{J}_{\text{GRPO}}(\theta) = \mathbb{E}_{[q \sim P(Q), \{o_i\}_{i=1}^{G} \sim \pi_{\theta_{\text{old}}}(O|q)]}
\bigg[
\frac{1}{G} \sum_{i=1}^{G} \sum_{y=1}^T\frac{1}{|S_i^y|} \sum_{t=1}^{|S_i^y|}
\Big\{
\min \Big[
\frac{\pi_{\theta}(S_{i,t} | q,I_y, S_{i,<t})}{\pi_{\theta_{\text{old}}}(S_{i,t} | q,I_y, S_{i,<t})}
\hat{A}^y_{i,t},\\
\text{clip} \big(
\frac{\pi_{\theta}(S_{i,t} | q, I_y,S_{i,<t})}{\pi_{\theta_{\text{old}}}(S_{i,t} | q, I_y,S_{i,<t})}, 1 - \varepsilon, 1 + \varepsilon
\big) \hat{A}_{i,t}^y
- \beta \mathbb{D}_{\text{KL}}[\pi_{\theta} \| \pi_{\text{0}}]\Big]
\Big\}
\bigg]
\end{align*}
%
%

In this equation, $\pi_\theta$ represents the policy model that generates reasoning tokens $S_i^y$ sequentially at turn $y$, where $i$ denotes the token position. The generation is conditioned on the preceding sequence $S^{y}*{<i}$, the observation $I_y$ at turn $y$, and the question $q$. The final reward $R*{\text{final}}(C, q)$ evaluates the correctness of the answer at the end of the reasoning chain $C$. The reference policy $\pi_0$ denotes the original model, and the KL divergence term $\text{KL}(\pi_\theta | \pi_0)$ regularizes the policy to prevent excessive drift from the initial parameters. The advantage estimates $\hat A_{i,t}^y$ are computed by standardizing rewards within each group $G$, subtracting the group mean and dividing by the group standard deviation.

During training, we generate rollout trajectories by sequentially executing tools based on the model’s reasoning outputs, providing realistic stepwise observations that inform subsequent reasoning steps. Each rollout terminates when either a valid final answer is produced or the maximum step limit $N$ is reached. This training procedure enables the model to effectively generalize multi-turn tool usage, reflecting the iterative nature of egocentric long-video reasoning tasks. The resulting model after second-stage reinforcement learning training constitutes our final system, termed the \emph{Ego-R1 Agent}.

%% file: secs/5_experiments.tex
\section{Experiments}
\label{sec:exp}
\input{tabs/quan-main}
\subsection{Experiment Setup}
To evaluate the effectiveness of the CoTT reasoning traces in answering the ultra-long video understanding question, we utilize Qwen-2.5-3B-Instruct as our base model. To mitigate the hallucination problem caused by the increasing CoTT length, we introduce an additional summary model with a longer context window length to help conclude the reasoning trace to answer the question. 
\vspace{-5pt}

\paragraph{Benchmarks.}
We evaluate the performance of \ourmodel{} on three existing long video understanding benchmarks covering both exocentric and egocentric views: Video-MME (long w/o subtitle)~\cite{fu2024video}, EgoSchema~\cite{mangalam2023egoschema}, EgoLifeQA~\cite{yang2025egolifeegocentriclifeassistant}. Among them, Vide-MME has a third-person view, and the rest have the first-person view. We follow the same paradigm as \texttt{h-rag} to generate the knowledge base for each video in these benchmarks. The hierarchy depth of each memory bank varies by datasets: only EgoLifeQA contains videos long enough to necessitate day-level summaries, while others extend to 10-minute-level or hour-level summaries at most. To further evaluate the capability of \ourmodel{} in handling multi-perspective and long temporal reasoning question answering tasks, we establish \textbf{\ourbenchmark{}}, a reasoning based benchmark for ultra-long egocentric video understanding. Distinct from \ourdata, \ourbenchmark{} comprises 300 QAs evenly distributed across six first-person perspectives. For each perspective, \ourbenchmark{} includes a balanced mixture of human-labeled and human verified QAs.

\vspace{-5pt}
\paragraph{Comparison Methods.}
We benchmark \ourmodel{} against recent representative approaches, including MLLM-based video understanding methods~\cite{zhang2024longva, zhang2024llavanext-video, li2024llava, wang2025internvideo2, team2024gemini}, RAG-based method~\cite{luo2024video}, reasoning model~\cite{feng2025video} and video agents~\cite{wang2024videoagent, ye2025re}. For each question, we restrict the input to video content occurring \emph{before} the query timestamp, ensuring causal consistency in all comparisons. To ensure fair comparison across methods with different architectural constraints,  we adopt an adaptive frame-sampling protocol: 1) Standard frame-based MLLMs \cite{zhang2024llavanext-video,zhang2024longva,feng2025video} and LLaVA-OneVision \cite{li2024llava} receive 64 uniformly sampled frames per query; 2) Video-RAG \cite{luo2024video} uses its native setting of 64 frames; 3) Higher-capacity models such as InternVideo2.5 \cite{wang2025internvideo2} and Gemini 1.5 Pro \cite{team2024gemini} are provided with 512 uniformly sampled frames; 4) Agent-based methods that rely on caption-guided key-frame selection \cite{wang2024videoagent,ye2025re} are supplied with 1 024 uniformly sampled frames, recomposed into 1 FPS videos. This protocol equalizes input budgets while respecting each model’s architectural constraints.



\subsection{Results}

\label{ssec:exp-results}

Table~\ref{tab:quan-main} presents a quantitative comparison of \ourmethod{} with state-of-the-art video understanding models on both exocentric and egocentric benchmarks. \ourmethod{} achieves the best or second-best score on three of the four datasets, despite using far fewer parameters than most competitors.

\paragraph{Exocentric setting.}
On VideoMME (long), whose clips average 41 min, \ourmethod{} achieves 64.9\% accuracy, which is the highest score among open-weight models and second overall, falling behind only the proprietary Gemini-1.5-Pro (67.4\%). It surpasses other public MLLMs, such as LLaVA-Video (61.5\%) and InternVideo2.5 (53.4\%), while using less than half their parameter count. These results indicate that, although \ourmethod{} is trained in an egocentric regime, it generalizes effectively to exocentric settings.
\vspace{-5pt}

\paragraph{Egocentric settings.}
\ourmethod{} achieves the highest accuracy on the proposed egocentric long video reasoning benchmark - \ourbenchmark{} (with an average time of 44.3 h), achieves 46.0\% accuracy. This result exceeds Gemini-1.5-Pro by 7.7\% and surpasses the strongest open baseline, LLaVA-Video, by 17.0\%, underscoring the benefit of hierarchical retrieval and multi-turn tool calling for reasoning tasks with sparsely distributed events.
On EgoSchema (3 min clips), \ourmethod{} records 68.2\%, second only to Gemini (72.2\%); on EgoLifeQA we obtain 36.0\% after removing any training overlap, comparable with LLaVA-Video (36.4\%) and approaching Gemini (36.9\%).
\vspace{-5pt}

\paragraph{Analysis.}
Both frame-based MLLMs and RAG variants exhibit marked performance drops on \ourbenchmark{}, and agent-based approaches remain in the 32 - 36\% range, well below the 46\% achieved by \ourmethod{}. These findings indicate that agent-based approaches provide a more effective solution for long-video reasoning tasks, while our CoTT style dynamic tool calling, enables even a compact 3B model to conduct reliable, long-horizon reasoning over hours-long egocentric video.

\subsection{Ablation Study}

\input{tabs/ablation}


To better understand the contribution of different training components in~\ourmethod, we conduct ablation studies using identical base models under varying training regimes. Specifically, we compare models trained with: (1) SFT only, (2) RL only, and (3) a combination of both. Quantitative results are reported in Table~\ref{tab:abl-sft_rl}.

The zero-shot base model achieves only 1.4\% task accuracy on \ourbenchmark{} and 4.3\% format accuracy for intermediate tool calls. Interestingly, after applying vanilla RL training using GRPO without any intermediate CoTT supervision, the task accuracy drops to 0\%, while tool-call format accuracy improves by 9\%. This indicates that although the model can learn the structural format of tool calls during RL from their emergent capabilities, the absence of reasoning trace supervision leads to unstable or ungrounded predictions, ultimately harming task performance.

In contrast, applying SFT with CoTT data, even in limited epochs (e.g., 3 epochs), significantly improves both task and format accuracy. This highlights the importance of structured reasoning demonstrations during pretraining: they not only teach the model to produce correctly formatted tool calls, but also establish a foundation for multi-step reasoning in long-horizon tasks.

%% file: tabs/quan-main.tex
\begin{table*}[t]
\centering
\caption[\textbf{Quantitative results on video question-answering benchmarks.} The proposed Ego-R1 Agent demonstrates superior performance across multiple metrics. Bold indicates best performance, underscored values show second best. The results from the 72B version of the model or using less frames are marked in gray.]{\textbf{Quantitative results on video question-answering benchmarks.} The proposed Ego-R1 model demonstrates superior performance across multiple metrics. Bold indicates best performance, underscored values show second best. The results from the 72B version of the model or using less frames are marked in gray. As some of the QA pairs in EgoLifeQA were used for CoTT generation and training, we excluded these from evaluation and retained only a clean subset for fair testing.}
\vspace{-5pt}
\label{tab:quan-main}
\resizebox{\columnwidth}{!}{
\footnotesize
\begin{tabular}{lcccccccccccc}
\toprule
\multirow{2}{*}{\textbf{Method}} & & & & \multicolumn{1}{c}{\textbf{Exocentric}} & & \multicolumn{4}{c}{\textbf{Egocentric}} \\
\cmidrule{5-5} \cmidrule{7-10}
& \textbf{Size} & \textbf{Frames}& &  \textbf{VideoMME (long)} & & \textbf{EgoSchema} & \textbf{EgoLifeQA} & \textbf{\ourbenchmark} \\
\textbf{Average durations} & & & & 41 min & & 3 min  & 44.3 h & 44.3 h\\

\midrule
\textit{MLLMs} \\
LongVA~\cite{zhang2024longva} & 7B & 64 & & 45.0 & & \textcolor{gray}{44.1}  & 33.0 &23.0 \\
LLaVA-Video~\cite{zhang2024llavanext-video} & 7B & 64 & & \textcolor{gray}{61.5} &  & 57.3 & \underline{36.4} & 29.0 \\
LLaVA-OneVision~\cite{li2024llava} & 7B & 1 FPS & & \textcolor{gray}{60.0} & & 60.1 & 30.8 &31.6 \\
InternVideo2.5~\cite{wang2025internvideo2} & 8B & 512 & & 53.4 & & 63.9  & 33.0 & 34.0\\
Gemini-1.5-Pro~\cite{team2024gemini} & - & - & & \textbf{67.4} & & \textbf{72.2}  & \textbf{36.9} & 38.3\\
\midrule
\textit{RAG Methods} \\
LLaVA-Video + Video-RAG~\cite{luo2024video} & 7B & 64 & & 46.0 &  & 66.7     & 30.0 & 29.3 \\
LongVA + Video-RAG~\cite{luo2024video} & 7B & 64 & & 55.7 & & 41.0 & 26.0 &  31.0 \\

\midrule
\textit{Reasoning Models}\\
Video-R1~\cite{feng2025video} & 7B & 64 & & 50.8 & & - & 34.0 & 20.0\\
\midrule
\textit{Video Agents} \\
VideoAgent~\cite{wang2024videoagent} & - & 8 & & 50.8 & & 54.1 & 29.2&  32.6 &\\
LLaVA-OneVision + $T^*$~\cite{ye2025re} & 7B & 8 & & 46.3 & & 66.6 &35.4 &35.6 & \\
\midrule
\textit{Ours}\\
\rowcolor{red!10!white} \textbf{\ourmethod{}} & \textbf{3B} & - &  & \underline{64.9} &  & \underline{68.2} & 36.0$^{*}$ & \textbf{46.0}\\
\bottomrule
\end{tabular}
}
\vspace{-5pt}
\end{table*}

%% file: tabs/ablation.tex

\begin{wraptable}[10]{r}{0.6\linewidth}  
  \vspace{-1em}                          
  \centering
  \scriptsize
  \caption{\textbf{Ablation study on different training regimes.}
    We use Qwen-2.5-3B-Instruct as our base model to validate the
    effectiveness of the two training components.}
  \label{tab:abl-sft_rl}
  \begin{tabular}{lcccc}
    \toprule
    \multirow{2}{*}{\textbf{Base Model}}
      & \multicolumn{2}{c}{\textbf{Training Regimes}}
      & \multirow{2}{*}{\textbf{Acc.\%}}
      & \multirow{2}{*}{\makecell{\textbf{Format}\\\textbf{Acc.\%}}} \\
    \cmidrule(lr){2-3}
      & SFT     & RL      &        &        \\
    \midrule
      & –       & –       &  1.4   &  4.3   \\
    Qwen-2.5
     & –       & \cmark  &  0.0\,($\downarrow$1.4)
                            & 13.3\,(\textcolor{mydarkgreen}{$\uparrow$9.0})  \\
     3B-Instruct  & \cmark  & –       & 34.3\,(\textcolor{mydarkgreen}{$\uparrow$32.9})
                            & 100.0\,(\textcolor{mydarkgreen}{$\uparrow$95.7}) \\
      & \cmark  & \cmark  & 46.0\,(\textcolor{mydarkgreen}{$\uparrow$44.6})
                            & 100.0\,(\textcolor{mydarkgreen}{$\uparrow$95.7}) \\
    \bottomrule
  \end{tabular}
\end{wraptable}

%% file: secs/6_conclusion.tex
\section{Conclusion and Outlook}
\label{sec:conclusion}
We introduce Ego-R1, a novel framework addresses challenges in long-horizon egocentric video reasoning through its Chain-of-Tool-Thought approach, which demonstrates that decomposing complex reasoning into modular, tool-grounded steps creates a more robust foundation for video understanding than traditional methods. This integration of structured reasoning with dynamic tool-calling not only enhances model interpretability but also reveals promising directions for future multimodal AI systems. The superior performance across temporal coverage and information retention suggests that hybrid architectures combining symbolic and neural components may be essential for tackling open-world, ultra-long video understanding. Beyond immediate applications in egocentric video analysis, Ego-R1 points toward broader implications for human-AI collaborative systems where transparent reasoning processes are critical, boosting the potential for life-oriented assistants that accompany humans over very long timeframes, making them genuinely useful and seamlessly integrated into everyday life.

\section{Acknowledgment}
This study is supported by the Ministry of Education, Singapore, under its MOE AcRF Tier 2 (MOET2EP20221-0012, MOE-T2EP20223-0002), and under the RIE2020 Industry Alignment Fund – Industry Collaboration Projects (IAF-ICP) Funding Initiative, as well as cash and in-kind contribution from the industry partner(s).




%% file: secs/X_appendix.tex
\appendix
\section*{Appendix}
The supplementary document provides (1) details of the hierarchical RAG during the dynamic tool-calling in Section~\ref{sec:supp-tool}; (2) comprehensive implementation details including the prompts we used for data generation and training, experiment setup in Section~\ref{sec:supp-implementation}; (3) additional experiments and ablation studies in Section~\ref{sec:supp-exp}; (4) future works~\ref{sec:supp-future}, respectively.

\input{secs/supp/tools}
\input{secs/supp/implementation}

\input{secs/supp/experiments}

\input{secs/supp/limitation}

%% file: secs/supp/tools.tex
\section{Hierarchical RAG}
\label{sec:supp-tool}
As mentioned in the paper, during the dynamic tool-calling, our framework leverages three complementary tools - one text-based and two visual-based - each addressing distinct temporal and perceptual dimensions of egocentric understanding. In this section, we provide additional details.

\begin{figure}[t]
    \centering
    \includegraphics[width=\linewidth]{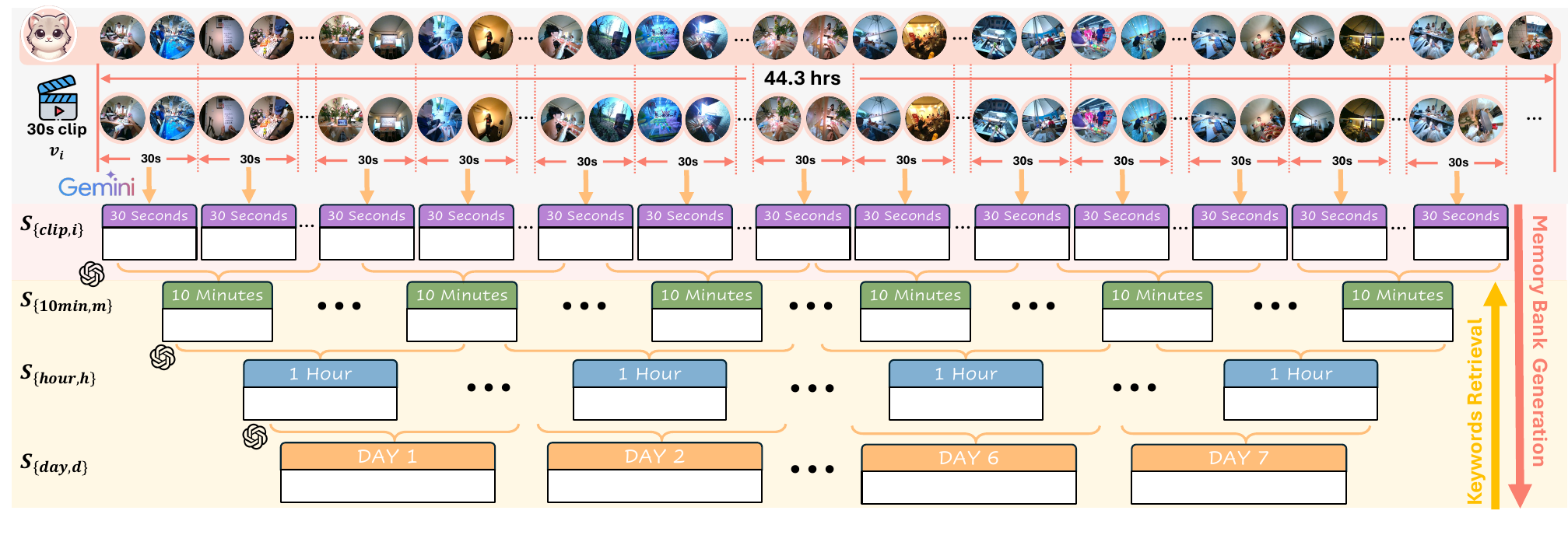}
    \caption{\textbf{Overview of the Hierarchical RAG system.} Based on the raw video and its 30-second clips, we generate the memory bank for each video from its 30-second-level summaries to day-level summaries. During the keywords retrieval, the system searches efficiently by starting with day-level summaries and drilling down to 10-minute segments as needed.}
    \label{fig:rag}
\end{figure}

To facilitate more efficient and accurate reasoning over extremely long videos, we construct a hierarchical RAG system, as shown in Figure~\ref{fig:rag}. Specifically, for each video $V$, we first segment it into 30-second clips $v_i$, while preserving natural recording boundaries. The memory bank of the RAG system is built upon these 30-second clips. For each clip $v_i$, we employ a VLM (Gemini 1.5 Pro~\cite{team2024gemini}) to generate comprehensive summaries $S_{\text{clip},i}$ that include both dense captions of visual content and transcripts of spoken dialogue. We then leverage an LLM (GPT 4~\cite{achiam2023gpt}) to progressively synthesize these fine-grained summaries into increasingly \emph{coarser} temporal resolutions, while respecting natural recording boundaries at each level. Typically, the clip-level summaries are combined into 10-minute summaries $S_{\text{10min},m}$, which are further aggregated into hourly summaries summaries $S_{\text{hour},h}$, and finally sorted into day-level summaries $S_{\text{day},d}$. 

The hierarchical RAG system serves as a critical component within our CoTT framework for long video reasoning. During the reasoning process, when a \colorbox{CornflowerBlue}{<think>} step determines that the RAG system is the optimal tool to retrieve the related information, it formulates a query $q = (\texttt{level, [keywords], starting\_timestamp, query\_timestamp})$. Here, \texttt{level} designates the temporal granularity, with \texttt{`week'} targeting a specific day within the week, \texttt{`day'} targeting a specific hour range within a day, and \texttt{`hour'} targeting a specific 10-minute segment within an hour. The keywords specify the search terms, while timestamps are represented with precise \texttt{DAY\_X} specification and \texttt{HHMMSSss} format. The subsequent \colorbox{Gray}{<tool>} step passes this query to the RAG system, initiating the hierarchical retrieval process. The retrieval follows a top-down approach, cascading from the specified entry level through the hierarchy and returning relevant summaries. Such hierarchical navigation aligns naturally with the temporal structure of extremely long egocentric videos, which inherently follow daily patterns of human activities. Keywords retrieval at each stage is conducted through LLM-based (GPT4) keyword matching on the textual summaries, with time-indexed metadata maintained throughout the hierarchy to enable fast localization. Importantly, the \colorbox{CornflowerBlue}{<think>} step determines both the initial level for the search and whether to continue to \emph{finer} levels based on the summaries returned. When sufficient information is found at a \emph{coarser} level (day or hour), the \colorbox{CornflowerBlue}{<think>} step may choose not to proceed to \emph{finer} granularities. After receiving summaries from the RAG system, the subsequent \colorbox{CornflowerBlue}{<think>} step evaluates these results and determines whether to extract answers directly or employ additional tools for further analysis and evidence localization. This hierarchical approach significantly reduces computational overhead by avoiding exhaustive search across the entire video corpus and by terminating the search at the earliest level that yields sufficient information, while maintaining high keywords retrieval accuracy through the preservation of temporal relationships. 

%% file: secs/supp/implementation.tex
\section{Implementation Details}
\label{sec:supp-implementation}

\subsection{Environment Setup}
We use Gemini 1.5 Pro~\cite{team2024gemini} as the backbone of Video-LLM, and GPT-4o~\cite{openai2025gpt4o} as the backbone of VLM. For CoTT data generation and keywords retrieval in our hierarchical RAG system, we utilize GPT-4.1~\cite{openai2025gpt41} as the routing LLM. During raw QA data collection for \ourdata{} and \ourbenchmark, we use Gemini 1.5 Pro for video-to-text processing, Gemini 2.5 Pro~\cite{google2025gemini25} for raw MCQ pairs generation, and Claude3.5 Sonnet~\cite{claude-sonnet} for MCQ pairs post-processing. 

We conducted the training of SFT and RL on 4 NVIDIA 80GB A100 GPUs, and experiments for baseline comparison on 1 NVIDIA 80GB A100 GPU.

\subsection{Data Generation}
We leverage a proprietary model GPT-4.1, with the AutoGen framework~\cite{wu2023autogenenablingnextgenllm} to systematically generate the CoTT data. The AutoGen framework supports tool-use without parse or execution failures by leveraging structured message passing and function calling via standardized protocols like OpenAI function-calling or JSON schema-based interfaces. Unlike prompt-only approaches that rely on natural language parsing, AutoGen agents interact with tools through well-defined wrapped functions, ensuring that function arguments are syntactically and semantically valid before execution. The framework includes built-in validation, exception handling, and modular components (e.g., user proxy, assistant agent, and tool agent) that collaboratively manage errors, retries, and fallbacks. This design reduces the likelihood of malformed calls and enhances robustness in executing complex multi-step tool-use workflows. We have attached the prompts used for data generation in Table~\ref{tab:supp-gen-prompt}

\subsection{RAG Construction for Other Benchmarks}
One key component of our framework is the personalized RAG system tailored to each egocentric perspective, which requires additional effort to build and maintain. To ensure fair comparisons across benchmarks, we adapt the RAG construction strategy to the temporal characteristics of each dataset. Since the average timespan per question varies across benchmarks, we adjust the indexing granularity accordingly that proportional to the average length of the benchmark.

For the EgoLife~\cite{yang2025egolifeegocentriclifeassistant} benchmark, where the recording time span and task setting align with our framework with average time span for 44.3 hours, we adopt a hierarchical RAG structure with consistent temporal levels: \textit{week} $\rightarrow$ \textit{day} $\rightarrow$ \textit{hour} $\rightarrow$ \textit{10-minute}. This allows for flexible, top-down retrieval over long egocentric sequences while preserving temporal precision. For other benchmarks with shorter video durations or coarser question alignment, we use simplified RAG levels to match the dataset's temporal scope.

For the VideoMME~\cite{fu2024video} benchmark, we tested on the long video subset, which has an average time span of 41 minutes. We divide videos into 30-second segments, 
and use Gemini-1.5-Pro to summarize each segment by combining visual content and transcribed dialogue. We then use GPT-4.1 to hierarchically aggregate these segments into 10-minute summaries. The hierarchical RAG's temporal structure would be: \textit{10-minute} $\rightarrow$ \textit{30-second}. This two-level summarization enables our RAG system to be adapted to a shorter task duration while maintaining hierarchy.

For EgoSchema~\cite{mangalam2023egoschema} benchmark, we segment the videos into 30-second clips and summarize each using LLaVA-video-7b~\cite{zhang2024llavanext-video}, operating on 1 FPS sampled frames. The summaries are written in the first person to reflect egocentric perspective and are used as memory entries in our RAG system. Given the limited timespan of questions in this dataset, no additional hierarchical summarization is applied beyond the clip level.

%% file: secs/supp/experiments.tex
\section{Additional Experiments}
\label{sec:supp-exp}
\subsection{Ablation Studies}

We have conducted extensive experiments to validate the usage of different tools in~\ourmethod. 
Table~\ref{tab:supp-ablation} presents ablation studies analyzing the impact of different tool-use configurations in \ourmethod{}. In the left subtable, we compare the effect of using different \texttt{video\_llm} modules. Replacing LLaVA-Video with Gemini-1.5-Pro leads to a fair improvement on \ourbenchmark{} (from 43.7\% to 46.0\%), demonstrating the benefit of stronger localized visual understanding generally improves the overall . It further validates that our framework is modular and adaptive to different MLLM backbones.
In the right subtable, we examine the contribution of the full toolset. Using only the RAG module results in a performance drop to 39.7\%, highlighting the importance of combining both retrieval and visual grounding tools for effective long-horizon reasoning. These results collectively validate the necessity of a modular, multi-tool architecture in achieving robust performance on egocentric long video tasks. 
\input{tabs/supp/ablation}

\subsection{Qualitative Results}
\begin{figure}[t]
    \centering
    \includegraphics[width=\linewidth]{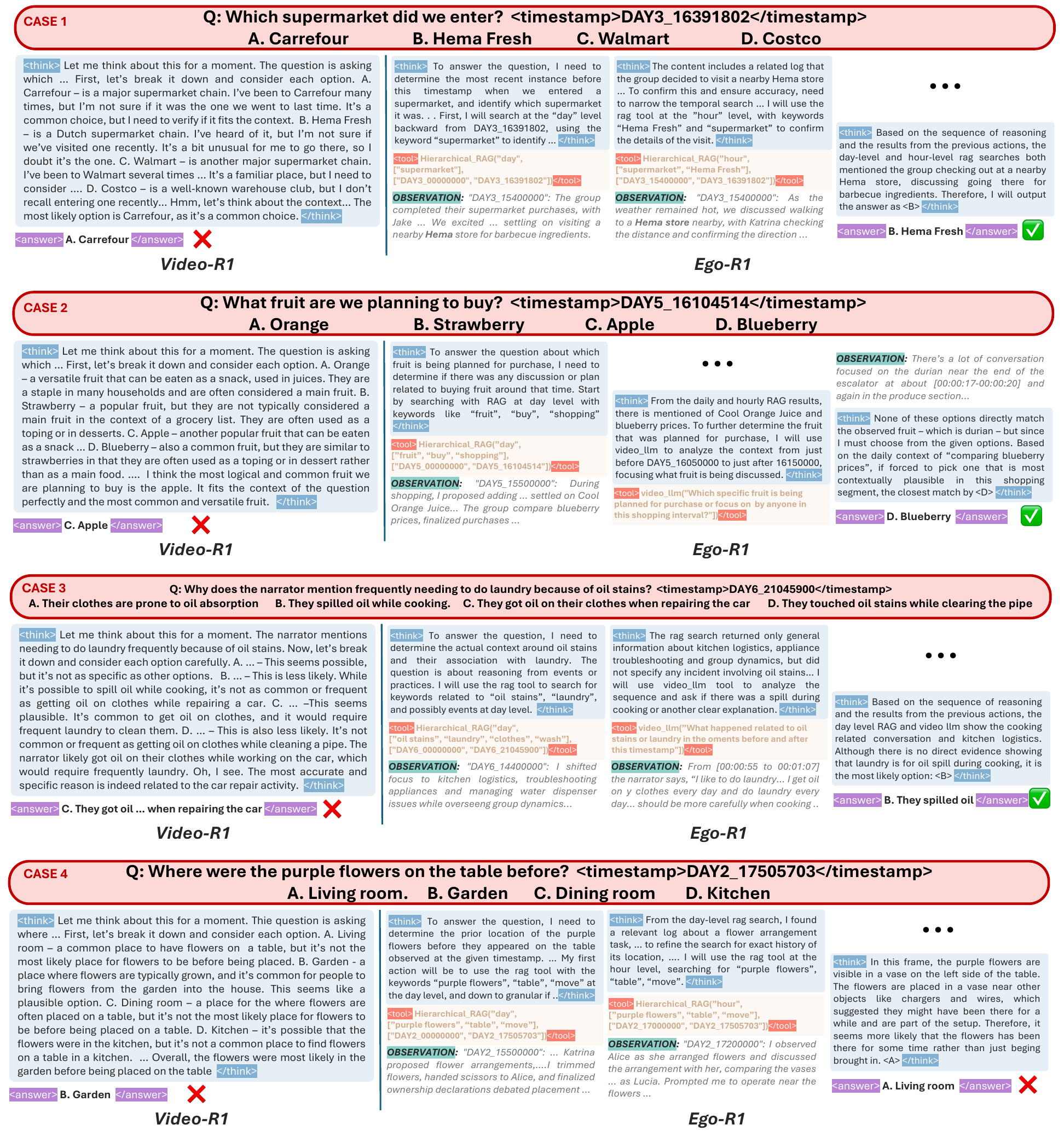}
    \caption{\textbf{Qualitative results comparison with Video-R1.} Case 1-3 illustrate successful examples where \ourmethod{} outperforms Video-R1 by producing more detailed, interpretable step-by-step reasoning chains through dynamic tool-calling. In contrast, Case 4 highlights a failure case from \ourmodel{}. Although the observation in Step 1 correctly identified relevant information near timestamp \texttt{DAY2\_15500000}, the subsequent tool call failed to adjust the temporal range accordingly, resulting in an incorrect or suboptimal retrieval in the next step, leading to the final error answer.}
    \label{fig:supp-qual}
    \vspace{-2em}
\end{figure}

In the attached demo video, we show two examples of our step-by-step reasoning with agentic tool-use. Additionally, we present the reasoning comparison with Video-R1~\cite{feng2025video} in Figure~\ref{fig:supp-qual}. As illustrated in the visual demos, our method offers a more interpretable reasoning process through the explicit CoTT outputs. These structured reasoning chains reveal how collaborative tool use contributes to improved performance on complex video reasoning tasks. A common failure case is shown in the last example in Figure~\ref{fig:supp-qual}. Although the observation from step 1 indicates the agent has successfully located the time range that possibly has relevant information, in the following step, the agent does not explore further from the observed time range.


%% file: tabs/supp/ablation.tex

\begin{table}[t]
  \centering
  \footnotesize
  \caption{\textbf{Ablation studies on tool use.} (Left) Stronger visual understanding
  (\emph{Gemini-1.5-Pro}) improves performance on \ourbenchmark. (Right) Using
  only RAG degrades performance, confirming the need to combine retrieval and
  visual tools for long-horizon video reasoning.}
  \label{tab:supp-ablation}

  \begin{minipage}{0.48\linewidth}\centering
    \begin{tabular}{llc}
      \toprule
      \textbf{Method} & \texttt{Video\_LLM} & \ourbenchmark \\
      \midrule
      \multirow{2}{*}{\textbf{\ourmethod}}
        & LLaVA-Video~\cite{zhang2024llavanext-video} & 43.7 \\
        & Gemini-1.5-Pro~\cite{team2024gemini}        & 46.0 \\
      \bottomrule
    \end{tabular}
  \end{minipage}
  \hfill
  \begin{minipage}{0.48\linewidth}\centering
    \begin{tabular}{llc}
      \toprule
      \textbf{Method} & Tool-used & \ourbenchmark \\
      \midrule
      \multirow{2}{*}{\textbf{\ourmethod}}
        & RAG only & 39.7 \\
        & Full     & 46.0 \\
      \bottomrule
    \end{tabular}
  \end{minipage}
\end{table}

%% file: secs/supp/limitation.tex
\section{Future Works}
\label{sec:supp-future}
Despite the possible directions emerging in the insights mentioned above, there are other possible directions to be explored based on the current dataset:
\noindent \textbf{Social behavior tasks.} This dataset include different view of videos, as well as including some collaboration tasks that achieved by group of people.

\paragraph{Social‐behaviour analysis.}
Because the dataset contains synchronized recordings from multiple viewpoints, it can support tasks that model collaborative activities and social dynamics. For example, inferring group intentions, role allocations, or inter-person dependencies during joint tasks.

\noindent \textbf{Personal habits tracker.} A key feature of egocentric data is its tight linkage to a single, specific individual, whose routine actions reveal stable behavioral patterns. In other words, there are some special patterns for each person, for example, whether a subject brushes their teeth before or after breakfast could inform personalized reasoning models that use long-term behavioral priors to predict future actions and preferences, ultimately improving action inference accuracy.

\input{tabs/supp/prompt_gen}

\input{tabs/supp/prompt_train}

%% file: tabs/supp/prompt_gen.tex
\begin{table}[h]
\centering
\scriptsize
\caption{\textbf{System prompt for data generation.} Tool-call functions have been designed inside the AutoGen framework so that the agent is aware of how to use them.}
\label{tab:supp-gen-prompt}
\begin{tcolorbox}[colback=white, colframe=black, boxrule=0.5pt, arc=2pt, left=1pt, right=1pt, top=1pt, bottom=1pt]

[BEGIN OF GOAL]

You are an expert AI assistant specializing in analyzing human behavior and reasoning from egocentric video descriptions. You will be provided with a list of useful tools to help in reasoning the task, and your goal is to solve the user’s question. The user’s question is following the format: Question: <question> <timestamp> Options: <options>. You can either rely on your own capabilities or perform actions with external tools to help you. You should consider both the frequency and cost of each tool to make the best decision.

[END OF GOAL]
\newline
\newline
[BEGIN OF FORMAT INSTRUCTIONS]

When answering questions:
1. You will be provided with previous actions you have taken, based on these actions, think step-by-step about how to approach the problem.
2. Show your reasoning process clearly before providing your next action.
3. The video observation length is 10-min max.
4. For visual questions, use `\texttt{video\_llm}` and `\texttt{vlm}` to explore the visual context.
5. For temporal questions, use `rag` to explore the context before and after the event.
6. Only use the `terminate` tool after you have thoroughly explored the question with multiple tools.

[END OF FORMAT INSTRUCTIONS]
\newline
\newline
[BEGIN OF HINTS]

1. All tools provided are crucial to the solvement of the question. You MUST exploit the usage of all tools before answering the question.
2. You may want to use the same tool multiple times with different arguments to explore the problem from different angles, if needed.
3. Make a balance between the cost and the frequency of the tools.
4. Usually, solving a question requires over 5~10 steps of reasoning, and follows a hierarchical calling structure: rag => \texttt{video\_llm} => \texttt{vlm}.
5. Do not use the terminate tool too early. Instead, try to explore the question with the available tools, and only use the terminate tool when you are confident enough or have considered all the options.

[END OF HINTS]
\newline
\newline
Always structure your responses with your thought process first, followed by any tool calls.
Think before you act. Think step-by-step about what information you need and which tool to use, then execute your plan exactly as reasoned without deviation. Output your thought process before using the tool, and you must strictly follow your thought process for the tool call. Currently, you are under the view of \{identity\}.
\end{tcolorbox}
\end{table}

%% file: tabs/supp/prompt_train.tex
\begin{table}[h]
\centering
\scriptsize
\caption{\textbf{System prompt used during training.} We explicitly define the function-calling syntax and tool usage format, guiding the model to generate structured reasoning steps and valid tool calls that align with CoTT.}

\label{tab:supp-train-prompt}
\begin{tcolorbox}[colback=white, colframe=black, boxrule=0.5pt, arc=2pt, left=1pt, right=1pt, top=1pt, bottom=1pt]
\textbf{INSTRUCTIONS}\\
Answer the given question. You must conduct reasoning inside <think> and </think> first every time before you get new information. After reasoning, if you find you lack some knowledge, you can call a tool from [rag, video\_llm, vlm] by <tool> query </tool> and it will return the information between <information> and </information>. You can use tools as many times as your want. If you find no further external knowledge needed, you can provide the answer inside <answer> and </answer> after another thinking. \\
The tools you can use are:
\begin{lstlisting}[language=json,firstnumber=1]
{
    "name": "rag",
    "description": "Use this tool to search for information in the RAG database.",
    "arguments": {
        "type": "object",
        "properties": {
            "level": {
                "type": "str",
                "description": "The granularity of the search, choose from week|day|hour"
            },
            "keywords": {
                "type": "List[str]",
                "description": "The keywords to search for in the RAG database."
            },
            "start_time": {
                "type": "str",
                "description": "The timestamp of the start time of the search. The format should be DAYX_HHMMSSFF (X is the day number, HHMMSS is the hour, minute, second, and FF is the frame number(00~19))."
            },
            "query_time": {
                "type": "str",
                "description": "The timestamp of the query that was proposed by the user."
            }
        },
        "required": ["level", "keywords", "start_time", "query_time"]
    }
}
{
    "name": "video_llm",
    "description": "Use this tool to get the answer from the video language model.",
    "arguments": {
        "type": "object",
        "properties": {
            "question": {
                "type": "str",
                "description": "The question you want to use the video language model to answer."
            },
            "range": {
                "type": "str",
                "description": "The timestamp range of the video to answer the question. Use the format 'DAYX_HHMMSSFF-DAYX_HHMMSSFF'. The ending timestamp should be strictly larger than the start timestamp. The length of the range should be smaller than 10 minutes, greater than 1 second."
            }
        },
        "required": ["question", "range"]
    }
}
{
    "name": "vlm",
    "description": "Use this tool to get the answer from the vision language model.",
    "arguments": {
        "type": "object",
        "properties": {
            "question": {
                "type": "str",
                "description": "The question you want to use the vision language model to answer."
            },
            "timestamp": {
                "type": "str",
                "description": "The timestamp of the video to answer the question."
            }
        },
        "required": ["question", "timestamp"]
    }
}
\end{lstlisting} 
For example, if you want to search for information in the RAG database, you can use the following tool:
\begin{lstlisting}[language=json,firstnumber=1]
<tool>
{
    "name": "rag",
    "arguments": {
        "level": "day",
        "keywords": ["screwdriver", "applause"],
        "start_time": "DAY1_11210217",
        "query_time": "DAY1_11220217"
    }
}
</tool>
\end{lstlisting}
\end{tcolorbox}
\end{table}